\documentclass{article}
\usepackage{fancyhdr}

\usepackage{microtype}
\usepackage{graphicx}
\usepackage{subcaption}
\usepackage{booktabs} 

\usepackage{hyperref}

\usepackage{graphicx}

\usepackage{amsmath}
\usepackage[table]{xcolor}

\usepackage{tikz}
\usepackage{fancyhdr}
\usepackage{tcolorbox}
\usepackage{multirow}
\usepackage{pgfplots}
\definecolor{lightgreen}{rgb}{0.88, 1, 0.88}
\definecolor{lightyellow}{rgb}{1.0, 1.0, 0.88}

\usepackage[preprint]{icml2026}
\makeatletter
\icml@noticeprintedtrue
\makeatother

\usepackage{amsmath}
\usepackage{amssymb}
\usepackage{mathtools}
\usepackage{amsthm}

\usepackage[capitalize,noabbrev]{cleveref}

\theoremstyle{plain}

\theoremstyle{definition}

\theoremstyle{remark}

\usepackage[textsize=tiny]{todonotes}

\title{TemporalGraphLLM: Temporal Graph Neural Networks with Large Language Models for Dynamic Text-Attributed Graphs}

\author{
Moran Beladev \quad
Or Eitan \quad
Gilad Katz \quad
Lior Rokach
\\[6pt]
\normalsize Department of Information Systems Engineering\\
\normalsize Ben-Gurion University of the Negev, Israel
\\[4pt]
\normalsize
\texttt{\{belachde,eitano,giladkz,liorrk\}@post.bgu.ac.il}\\
}

\date{}
\icmltitlerunning{TemporalGraphLLM: Temporal Graph Neural Networks with Large
Language Models for Dynamic Text-Attributed Graphs}
\begin{document}

\maketitle
\begin{abstract}
Dynamic text-attributed graphs (DTAGs), where nodes, edges, and textual attributes evolve over time, are crucial in applications such as social networks, citation graphs, and knowledge graphs. However, existing approaches struggle to jointly model the temporal evolution of graph structures and the semantic richness of textual attributes. While Temporal Graph Neural Networks (TGNNs) capture evolving node relationships, they often lack contextual text reasoning. Conversely, Large Language Models (LLMs) excel in textual understanding but struggle with structured graph reasoning in temporal settings. To bridge this gap, we propose \textbf{TemporalGraphLLM}, a novel framework that can integrate any temporal GNN with an LLM for enhanced reasoning in DTAGs. Our approach fine-tunes LLMs using graph-time-aware instruction tuning and novel temporal GNNs injection to replace dedicated added tokens with graph embeddings. TemporalGraphLLM effectively leverages pretrained TGNNs within an LLM framework to achieve state-of-the-art performance on edge classification, link prediction, and edge-based text generation tasks. Extensive evaluation on real-world dynamic graph datasets demonstrates state-of-the-art performance. Our findings highlight the synergistic potential of LLMs and TGNNs, opening new directions for learning on evolving graphs.
\end{abstract}

\section{Introduction}
Graphs are a fundamental data structure for modeling relationships in various domains, including social networks, knowledge graphs, and citation networks. Dynamic Text-Attributed Graphs (DTAGs) represent entities and their evolving relationships over time, with both nodes and edges carrying textual attributes. Unlike static graphs, DTAGs pose unique challenges due to their temporal evolution and rich textual nature.

\begin{figure}[ht!]
    \centering
    \includegraphics[width=0.5\textwidth]{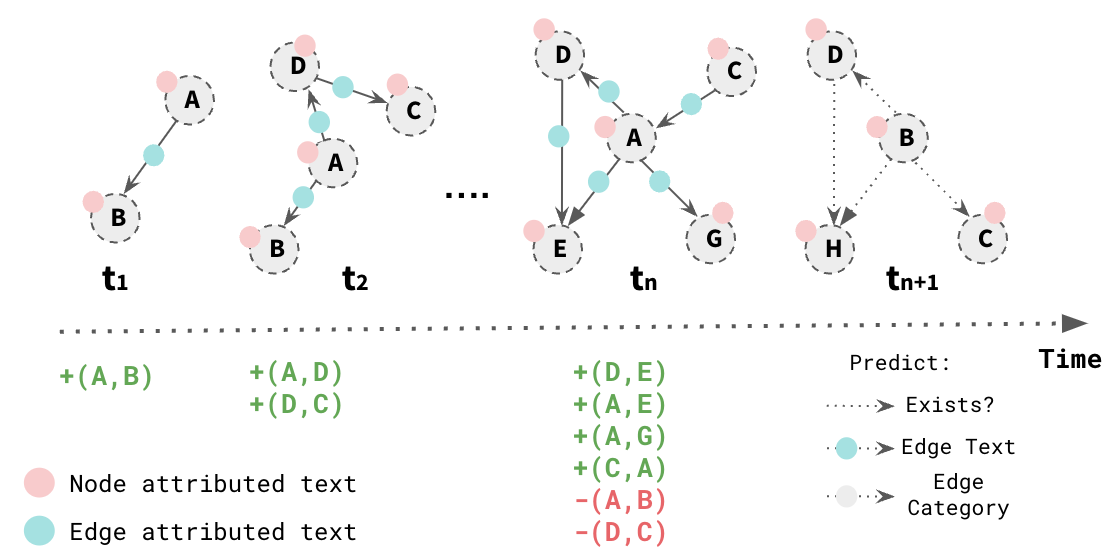}
    \caption{A Dynamic Text-Attributed Graph (DTAG) example, where nodes and edges evolve over time with textual attributes. Our work aims to enhance predictive capabilities over such a structure.}
    \label{fig:DTAG_example}
\end{figure}

As illustrated in Figure~\ref{fig:DTAG_example}, nodes (e.g., entities) and edges (e.g., relationships) dynamically change over discrete time steps. The textual attributes associated with nodes and edges evolve, while new connections emerge, and old ones disappear. This dynamic nature makes tasks such as link prediction, event forecasting, and textual relationship generation significantly more complex compared to traditional graph learning. For example, in a user-product graph, a textual relationship might be a user review. Generating such time-sensitive and context-aware text requires models to reason jointly over structural, textual, and temporal signals.


Existing graph neural network (GNN) approaches, including Temporal Graph Neural Networks (TGNNs) such as TGAT \cite{tgat2019}, DyGFormer \cite{dygformer2023}, and GraphMixer \cite{graphmixer2023}, capture evolving node interactions but struggle with textual reasoning and long-range dependencies. While effective at capturing structural dynamics, these TGNNs typically do not inherently process or leverage rich textual attributes associated with nodes and edges, a gap LLMs can fill. Conversely, LLMs excel in semantic understanding but lack relational inductive biases to process structured graph data effectively \cite{Li2024}. Current methods for LLM-GNN fusion typically fall into two categories:
\textbf{LLM as Enhancer}: Augmenting GNN node representations using LLM-derived embeddings or explanations; and, \textbf{LLM as Predictor}: Using LLMs to make graph-based predictions via textual reformulation of graph data. Despite these advances, a key gap remains in integrating temporal node representations with LLM-based reasoning. This gap is particularly large for evolving graphs where nodes, edges, and textual attributes change over time.

To bridge this gap, we introduce \textbf{TemporalGraphLLM}, a novel framework that seamlessly integrates temporal GNNs with LLMs to enable reasoning over dynamic text-attributed graphs. Specifically, our contributions are:
\begin{itemize}
\item We propose a generalizable TGNN--LLM architecture for dynamic text-attributed graphs that pairs \emph{any} TGNN with \emph{any} LLM via a temporal injection interface. Specifically, we \emph{replace} dedicated placeholder token representations inside the LLM stream with time-aware TGNN embeddings through learned projections, aligning graph structure, text, and temporal evolution.

\item We enable end-to-end \emph{joint} training of the TGNN and LLM under temporally-aware instruction tuning. Our training pipeline supports temporal neighbor sampling for DyTAGs and optimizes a multi-objective loss that couples structural/temporal representation learning with text generation quality, allowing gradients to flow through the projection into the TGNN.



\item We demonstrate the effectiveness of TemporalGraphLLM through extensive benchmarking on email networks, social graphs, e-commerce graphs, and evolving knowledge graphs. Our model significantly outperforms existing approaches in temporal edge classification, link prediction, and edge text generation.
\item Finally, we make the code publicly available.
\end{itemize}

\section{Related Work}

Recent years have seen growing research on combining LLMs and GNNs to enhance reasoning on graph-structured data. \citet{Li2024} outline three main approaches—\emph{LLM as Enhancer}, \emph{LLM as Predictor}, and \emph{LLM as Alignment Component}—covering strategies that leverage both textual and structural cues in graph learning. The challenge of incorporating \emph{temporal} reasoning into LLM–GNN frameworks remains largely unexplored; we address this gap by integrating TGNNs with LLMs for dynamic, text-rich graphs.

\paragraph{(1) \textbf{LLM as Enhancer}.}
LLMs can enrich GNNs on text-attributed graphs by: (a) supplying contextualised text embeddings as initial node features (e.g., \citeauthor{llmenhancer}) and (b) providing textual explanations or semantic insights that enrich node representation and aid in downstream tasks, as in TAPE~\cite{tape2024} and the LLM-to-LM Interpreter~\cite{he2024harnessing}.
\paragraph{(2) \textbf{LLM as Predictor}.} LLMs have also been used to make predictions on graph-based problems:
\textit{Flatten-Based Prediction} serializes graph structure as sequences for language modeling, with models like InstructGLM \cite{instructglm2024}, GPT4Graph \cite{gpt4graph}, and GraphText \cite{graphtext} reformulating graph data as textual input to exploit LLMs' capabilities.
\textit{GNN-Based Prediction} combines GNN-extracted structural embeddings with LLMs through techniques like prefix tuning or task-specific alignment, as in GraphLLM \cite{graphllm}, GraphGPT \cite{tang2024graphgpt}, and DGTL \cite{dgtl}.
\paragraph{(3) \textbf{LLM as Alignment Component}.}
Properly aligning GNN and LLM embedding spaces is crucial for integrating structured and unstructured data. \textit{Contrastive learning} technique maintains consistency between graph and text representations \cite{radford2021learning, liu2023multi, su2022molecular};
\textit{Iterative or graph-nested alignment} (e.g., GLEM \cite{glem} and GraphFormers \cite{yang2021graphformers}) that refine the fusion through multiple feedback steps or combine GNNs with Transformers for unified representations;
\textit{Distillation-based alignment}, where GNNs teach LLMs graph-aware reasoning\cite{mavromatis2023train}.

\paragraph{LLMs for Dynamic Graphs \& Temporal KGs.}
Work on evolving graphs is more recent.  LKD4DyTAG distils edge-level semantics from an LLM into a TGNN, leaving the LLM frozen~\cite{roy2025lkd4dytag}.  LLM-DA extracts temporal rules with GPT-4 and updates them as facts change~\cite{wang2024llmda}.  \textit{Back to the Future} introduces the \emph{ExpTime} benchmark and trains TimeLlama for explainable forecasting~\cite{yuan2023future}.  TimeR$^{4}$ addresses temporal KG question answering via a retrieve–rewrite–rerank loop~\cite{qian-etal-2024-timer4}.
Unlike the above, we \emph{co-train} a temporal GNN with an LLM, injecting time-sensitive node embeddings directly into the LLM and encouraging a shared latent interface via projection and joint supervision.

\paragraph{Temporal GNN Baselines.}
TGNNs such as TGAT~\cite{tgat2019}, DyGFormer~\cite{dygformer2023}, GraphMixer~\cite{graphmixer2023}, tdGraphEmbed~\cite{beladev2020tdgraphembed} and GraphERT~\cite{beladev2023graphert} capture temporal patterns with attention- or Transformer-style mechanisms but do not leverage LLM reasoning.  DTGB~\cite{zhang2024dtgb} further shows that most text-attributed graph methods ignore time, and that naively appending LLM embeddings offers limited benefit.

\paragraph{Our Contribution.}
We bridge these gaps by \emph{jointly training} a TGNN and an LLM with a dedicated temporal injection interface: the TGNN produces time-aware node embeddings that \emph{replace} placeholder token representations in the LLM stream via learned projections. We co-train both modules end-to-end with a multi-objective loss, unifying structural, temporal, and textual supervision and yielding stronger predictive and generative performance on dynamic, text-attributed graphs.

\section{The Proposed Approach: TemporalGraphLLM}

\subsection{Dynamic Text-Attributed Graphs -- Notations} \label{tgnn}

Let $G$ be a dynamic text-attributed graph, where $V$ and $E$ denote the node and edge sets, respectively. A temporal subgraph $G_T = (V_T, E_T)$ contains all nodes and edges observed up to timestamp $T$. Each node $v \in V_T$ is associated with a textual description $d_v \in D$, and each edge $(u, v) \in E_T$ is associated with a textual relation $r_{u,v} \in R$, category label $l_{u,v} \in L$, and timestamp $t_{u,v} \in T$. We denote the set of neighbors of node $v$ at time $t$ by $\mathcal{N}_v(t)$.

The goal of a Temporal Graph Neural Network (TGNN) is to learn time-aware node embeddings $h_v^t$ that capture both structural and temporal dependencies that are then fed to later stages of TemporalGraphLLM.

\subsection{TemporalGraphLLM}
\begin{figure*}[ht!]
    \centering
    \includegraphics[width=0.85\textwidth]{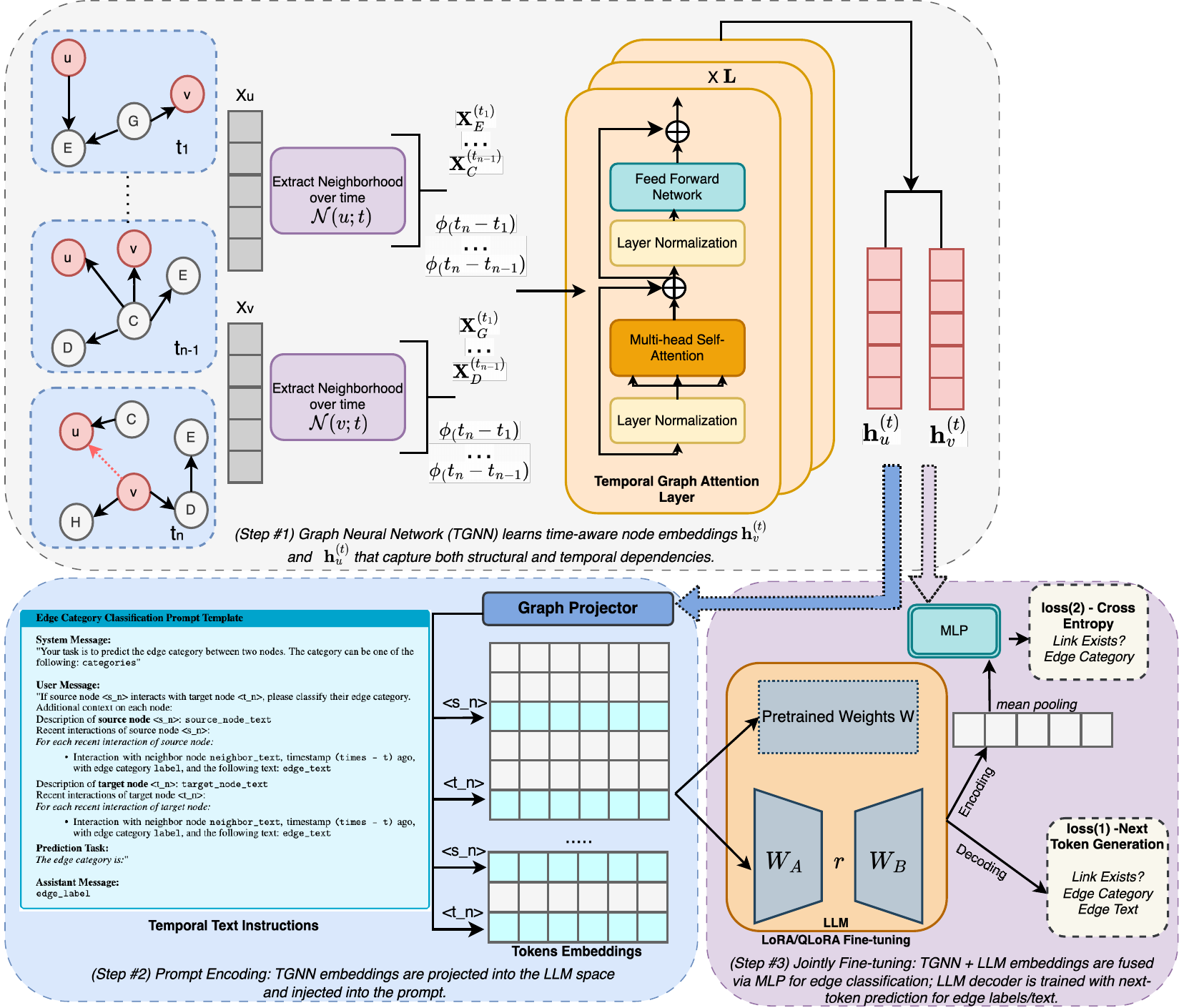}
    \caption{Overview of the TemporalGraphLLM architecture. (Step \#1) A TGNN computes time-aware node embeddings by incorporating temporal neighborhood information. (Step \#2) These embeddings are projected into the LLM space and injected into a prompt alongside node/edge text in temporal context. (Step \#3) During fine-tuning, TGNN and LLM embeddings are fused for edge prediction using an MLP, and the LLM decoder is trained via next-token prediction.}
    \label{fig:DTAG}
\end{figure*}

We propose \textbf{TemporalGraphLLM}, a novel framework that integrates TGNNs with Large Language Models (LLMs) to enable reasoning over Dynamic Text-Attributed Graphs. By fusing structural, temporal, and textual representations, our approach empowers LLMs to incorporate rich temporal graph embeddings for a variety of downstream tasks---namely, edge classification, link prediction, and edge text generation. 
The framework comprises the following three stages (illustrated in Figure~\ref{fig:DTAG}): \\
\textbf{(Step \#1) Temporal Graph Embedding:} We apply TGNN (e.g., TGAT, GraphMixer, or DyGFormer) to compute time-aware node embeddings $h_u^t$ and $h_v^t$ for each edge $(u, v)$ based on the graph structure and historical interactions.\\
\textbf{(Step \#2) Prompt Encoding:} We construct a textual prompt that includes the edge’s node and relation descriptions, and inject the temporal graph embeddings (from step \#1) by replacing special placeholder tokens with their projected representations in the LLM’s embedding space. This enables the LLM to incorporate both textual and temporal-structural context. \\
\textbf{(Step \#3) Fine-tuning and Prediction:} The LLM and TGNN are jointly fine-tuned on downstream tasks. We supervise the LLM with a language-modeling loss over the label text: for classification, each class corresponds to a dedicated label token (single-token target), and when using a textual description we compute cross-entropy over the sequence of tokens that realizes that description. For classification tasks (link prediction and edge classification), we additionally apply a classification loss by aggregating the LLM’s last hidden states and fusing them with the TGNN embeddings via a multi-layer perceptron (MLP). This dual supervision enables the LLM to act both as an \textit{encoder} (integrating graph and text context) and a \textit{decoder} (producing predictions or generating edge text). 

In the following sections, we review each step in detail.


\subsubsection{Temporal GNN Embedding}
Given a dynamic graph $G_T = (V_T, E_T, D, R, L, T)$, we can use any temporal GNN to obtain time-aware node embeddings.
For a given edge $(u, v)$ occurring at timestamp $t_{u,v}$, we compute node embeddings using the temporal GNN:
\begin{equation}
\mathbf{h}_u^t, \mathbf{h}_v^t = \text{TemporalGNN}(u, v, t_{u,v})
\end{equation}
Step \#1 in Figure \ref{fig:DTAG}  illustrates the general architecture of TGNN. 
Given any TGNN backbone and a query triple \((u,v,t)\),
we first gather each node’s \emph{time-ordered interaction history}
\(\mathcal{N}_u(t)\) and \(\mathcal{N}_v(t)\), i.e.\ all edges incident
to \(u\) or \(v\) that occurred before time \(t\).
The initial feature embeddings for nodes $u$ and $v$, denoted as $X_u$ and $X_v$, are derived from LLM embeddings of their textual descriptions (followed by LLM-as-Enhancer approaches \cite{llmenhancer}).

\noindent\textit{Time encoding.} We adopt a generic TGNN backbone that processes time-ordered interaction sequences for the queried nodes. When explicit time features are used, we append a continuous-time encoding $\phi(\Delta t)$, with $\Delta t = t - t_i$, to each event, where $\phi(\cdot)$ maps elapsed time to a feature vector (e.g., sinusoidal/Fourier or a learned variant). The resulting time-augmented sequences are passed through the backbone’s message-passing or token-mixing layers, stacked $L$ times, yielding time-aware node representations $\mathbf{h}^t_u$ and $\mathbf{h}^t_v$ that capture temporal dynamics and multi-hop structure. 

\noindent We instantiate this generic template with representative backbones —\textbf{TGAT}~\cite{tgat2019} (graph attention with continuous-time positional encodings), \textbf{GraphMixer}~\cite{graphmixer2023} (MLP-only mixer combining temporal and structural tokens for scalability), and \textbf{DyGFormer}~\cite{dygformer2023} (memory-augmented Transformer capturing long-range dependencies via co-occurrence patching)—but the method is agnostic to the specific choice. All three fit naturally into this pipeline and feed their learned embeddings to later stages of TemporalGraphLLM.

\subsubsection{Prompt Encoding}

\noindent\textbf{Graph projector.}
We project TGNN node embeddings into the LLM embedding space via learned linear maps (see Graph Projector block under step \#2 in Figure \ref{fig:DTAG}):  
\begin{equation}
graph  \: projector: 
\mathbf{e}_u = W_u \mathbf{h}_u^t, \quad \mathbf{e}_v = W_v \mathbf{h}_v^t
\end{equation}
where $W_u,W_v \in \mathbb{R}^{d_\ell \times d_g}$ map $d_g$-dimensional TGNN vectors to the LLM token space $d_\ell$.

\noindent\textbf{Temporal neighbor sampling.}
To improve the LLM’s understanding of temporal context, we introduce temporal-based prompt instructions by incorporating sampled temporal neighbors. 
Given the set of past neighbors $\mathcal{N}_u(t)$ for a node $u$, we sample a fixed number $k$ of neighbors using a temporal probability distribution: 
\begin{equation} 
\label{eq:temporal_sampling}
P(n_i) = \frac{t_i + 1}{\sum_{j} (t_j + 1)}, 
\end{equation} 
where $P(n_i)$ is the sampling probability for neighbor $n_i$ with timestamp $t_i$, and $t_j$ denotes the timestamps of all neighbors in $\mathcal{N}_u(t)$. 
All neighbor timestamps satisfy $t_i,t_j < t$ and lie within the dataset’s discrete timeline (from the start of observations up to the query time $t$).
This recency-weighted distribution favors more recent interactions (larger $t_i$) while ensuring older ones retain nonzero probability via the $+1$ term, which empirically improved temporal consistency in our prompts.
The effect of the neighbor count $k$ on model performance is systematically analyzed in the ablation study (see Section~\ref{sec:ablation}).
A comparison between temporal and random neighbor sampling, provided in Appendix~\ref{ablation_appendix}, further demonstrates that recency-aware sampling yields consistently higher performance metrics across all TGNN backbones.

\noindent\textbf{Prompt construction and injection.}
We add two special tokens, \texttt{<s\_n>} and \texttt{<t\_n>}, to the tokenizer.
From the edge’s local context (its node and edge texts, sampled temporal neighbors) we form a prompt $\mathbf{X}_{\text{orig}}$ containing these placeholders. Prompts are generated automatically from a single reusable template whose placeholders (e.g., \{node\_text\}, \{neighbor\_descriptions\}, \{relation\_type\}) are filled from dataset metadata and sampled histories  (see full prompt in Appendix \ref{prompts}), with no manual tuning.

Immediately before the Transformer blocks, we \emph{override} the special tokens embedding activations with the projected vectors $\mathbf{e}_u,\mathbf{e}_v$ and feed the sequence to the LLM:
\begin{equation}
\mathbf{X}_{\text{LLM}}=\textsc{ReplaceTokens}\!\left(\mathbf{X}_{\text{orig}},\mathbf{e}_u,\mathbf{e}_v\right).
\end{equation}
This operation overrides the placeholder tokens’ embedding activations inside the LLM stream, while preserving positional indices, so absolute or rotary position encodings are applied exactly as in the base model. Gradients at the replaced positions flow through $W_u,W_v$ into the TGNN, enabling end-to-end joint optimization.

We use $S{=}2$ injection slots by default (\texttt{<s\_n>}, \texttt{<t\_n>}), and the mechanism extends naturally to $S>2$.

\subsubsection{Edge Prediction via LLM-GNN Fine-tuning}
As illustrated in step \#3 in Figure \ref{fig:DTAG}, the LLM and TGNN are jointly fine-tuned to support multiple downstream tasks. 
The hidden states from the LLM are aggregated using mean pooling to produce a contextual embedding:
\begin{equation}
\mathbf{h}_{\text{LLM}} = \text{meanpool}(\text{LLM}(\mathbf{X}_{\text{LLM}}))
\end{equation}
For both \textit{edge classification} and \textit{link prediction}, we combine the temporal GNN node embeddings with the LLM-derived context using a multi-layer perceptron (MLP):
\begin{align}
\mathbf{z} &= \sigma(W_1 [\mathbf{h}_u^t || \mathbf{h}_v^t] + b_1) \\
\mathbf{z'} &= \sigma(W_2 [\mathbf{z} || \mathbf{h}_{\text{LLM}}] + b_2) \\
\hat{\mathbf{y}} &= f(W_3 \mathbf{z'} + b_3)
\end{align}
where \( || \) denotes vector concatenation, \( \sigma \) is a non-linear activation (e.g., ReLU), and \( f \) is the final activation:
$f = \text{sigmoid}$ for binary link prediction,
$f = \text{softmax}$ for multi-class edge classification.

We define the \textbf{edge-level loss} \( \mathcal{L}_{\text{edge}} \) based on the prediction task: binary cross-entropy for link prediction, and multi-class cross-entropy for edge classification.

\begin{table*}[t]
    \centering
    \caption{Statistics of datasets and comparison with existing datasets.}
    \label{tab:dataset_stats}
    \small
    \setlength{\tabcolsep}{1.8pt} 
    \renewcommand{\arraystretch}{1.2} %
    \begin{tabular}{cccccccc}
        \hline
        \multicolumn{1}{c}{\textbf{Dataset}} & \multicolumn{1}{c}{\textbf{Nodes}} & \multicolumn{1}{c}{\textbf{Edges}} & \multicolumn{1}{c}{\textbf{Edge}} & \multicolumn{1}{c}{\textbf{Timestamps}} & \multicolumn{1}{c}{\textbf{Domain}} & \multicolumn{1}{c}{\textbf{Text}} & \multicolumn{1}{c}{\textbf{Bipartite}} \\
        & & & \multicolumn{1}{c}{\textbf{Categories}} & & & \multicolumn{1}{c}{\textbf{Attributes}} & \multicolumn{1}{c}{\textbf{Graph}} \\
        \hline
        Enron & 42,711 & 797,907 & 10 & 1,006 & E-mail & Node \& Edge & No \\
        GDELT & 6,786 & 1,339,245 & 237 & 2,591 & Knowledge graph & Node \& Edge & No \\
        Stack elec & 397,702 & 1,262,225 & 2 & 5,224 & Multi-round dialogue & Node \& Edge & Yes \\
        Googlemap CT & 111,168 & 1,380,623 & 5 & 55,521 & E-commerce & Node \& Edge & Yes \\
        Amazon movies & 293,566 & 3,217,324 & 5 & 7,287 & E-commerce & Node \& Edge & Yes \\
        \hline
    \end{tabular}
\end{table*}

\begin{table*}[t]
\caption{Results for the Link Prediction task.}
\label{link_prediction_table}
\centering
\scriptsize
\setlength{\tabcolsep}{3pt}
\renewcommand{\arraystretch}{1.2}
\begin{tabular}{|c|c|c|cc|cc|cc|cc|cc|}
\hline
\textbf{Method} & \textbf{Graph-Model} & \textbf{LLM} 
& \multicolumn{2}{c|}{\textbf{Enron}} 
& \multicolumn{2}{c|}{\textbf{Googlemap CT}} 
& \multicolumn{2}{c|}{\textbf{Stack Elec}}
& \multicolumn{2}{c|}{\textbf{GDELT}} 
& \multicolumn{2}{c|}{\textbf{Amazon Movies}} \\
\cline{4-13}
& & & \textbf{Acc.} & \textbf{F1} 
& \textbf{Acc.} & \textbf{F1} 
& \textbf{Acc.} & \textbf{F1} 
& \textbf{Acc.} & \textbf{F1} 
& \textbf{Acc.} & \textbf{F1} \\
\hline

\multirow{3}{*}{\parbox{1.5cm}{\centering \textbf{Graph-based}\\\textbf{methods}}}
& DyGFormer & -
& 0.7390 & 0.7177 & \cellcolor{lightyellow}0.8336 & 0.7919 & 0.7983 & 0.8353 & 0.7583 & 0.7529 & 0.8293 & 0.8103 \\
& GraphMixer & -
& 0.7270 & 0.7083 & 0.8436 & 0.8102 & 0.7833 & 0.8239 & 0.7541 & 0.7652 & 0.8190 & 0.8249 \\
& TGAT & & 0.7225 & 0.7220 & 0.8276 & 0.8081 & 0.7858 &  \cellcolor{lightyellow}0.8436 & 0.7539 & 0.7603 & 0.8245 & 0.8267 \\
\hline
\hline

\multirow{3}{*}{\parbox{1.5cm}{\centering \textbf{LLM-based}\\\textbf{methods}}}
& - & Llama3-1B 
& 0.7085 & 0.7354 & 0.7055 & 0.6549 & 0.7019 & 0.7352 & 0.6695 & 0.6671 & 0.7277 & 0.7494 \\
& - & Llama3-3B 
& 0.6792 & 0.7230 & 0.7078 & 0.6450 & 0.7146 & 0.7318 & 0.6470 & 0.6451 & 0.7417 & 0.7468 \\
& - & Mistral-7b 
 & 0.6882 & 0.7416 & 0.7181 & 0.6447 & 0.7038 & 0.7267 & 0.6538 & 0.6525 & 0.7513 & 0.7668 \\
\hline
\hline

\multirow{9}{*}{\parbox{1.5cm}{\centering \textbf{Ours – Temporal}\\\textbf{Graph LLM}}}
& DyGFormer & Llama3-1B 
& \cellcolor{lightgreen}0.7512 & 0.7345 & 0.7842 & \cellcolor{lightgreen}0.7984 & \cellcolor{lightgreen}0.8863 & \cellcolor{lightgreen}0.8871 & 0.7534 & \cellcolor{lightgreen}0.7883 & 0.8238 & \cellcolor{lightgreen}0.8680 \\
& & Llama3-3B 
& \cellcolor{lightgreen}0.7484 & \cellcolor{lightgreen}0.7232 & 0.8241 & \cellcolor{lightgreen}0.8391 & \cellcolor{lightgreen}0.8849 & \cellcolor{lightgreen}0.8620 & \cellcolor{lightgreen}0.7834 & \cellcolor{lightgreen}0.7805 & \cellcolor{lightgreen}0.8380 & \cellcolor{lightgreen}0.8537 \\
& & Mistral-7b 
& \cellcolor{lightgreen}0.7600 & 0.7279 & 0.8180 & \cellcolor{lightgreen}0.8356 & \cellcolor{lightgreen}\textbf{0.9053} & \cellcolor{lightgreen}\textbf{0.9090} & \cellcolor{lightgreen}0.7848 & \cellcolor{lightgreen}0.7938 & \cellcolor{lightgreen}0.8313 & \cellcolor{lightgreen}0.8509 \\

& GraphMixer & Llama3-1B 
& \cellcolor{lightgreen}0.7484 & 0.7221 & \cellcolor{lightgreen}0.8645 & \cellcolor{lightgreen}\textbf{0.8770} & \cellcolor{lightgreen}0.8861 & \cellcolor{lightgreen}0.8779 & 0.7501 & \cellcolor{lightgreen}0.7677 & 0.8103 & 0.8152 \\
& & Llama3-3B 
& \cellcolor{lightgreen}0.7387 & \cellcolor{lightgreen}0.7294 & \cellcolor{lightgreen}0.8614 & \cellcolor{lightgreen}0.8687 & \cellcolor{lightgreen}0.8901 & \cellcolor{lightgreen}0.8881 & \cellcolor{lightgreen}0.7597 & 0.7561 & \cellcolor{lightgreen}0.8264 & \cellcolor{lightgreen}0.8326 \\
& & Mistral-7b 
& \cellcolor{lightgreen}0.7418 & \cellcolor{lightgreen}0.7495 & \cellcolor{lightgreen}\textbf{0.8742} & \cellcolor{lightgreen}0.8755 & \cellcolor{lightgreen}0.8941 & \cellcolor{lightgreen}0.8890 & \cellcolor{lightgreen}0.7771 & 0.7629 & \cellcolor{lightgreen}0.8349 & \cellcolor{lightgreen}0.8415 \\

& TGAT & Llama3-1B 
& \cellcolor{lightgreen}0.7654 & \cellcolor{lightgreen}0.7587 & \cellcolor{lightgreen}0.8494 & \cellcolor{lightgreen}0.8594 & \cellcolor{lightgreen}0.8449 & 0.8310 & \cellcolor{lightgreen}0.7776  & \cellcolor{lightgreen}0.7818 & \cellcolor{lightgreen}0.8498 & \cellcolor{lightgreen}\textbf{0.8687} \\
& & Llama3-3B 
& \cellcolor{lightgreen}0.7631 & \cellcolor{lightgreen}0.7657 & \cellcolor{lightgreen}0.8590 & \cellcolor{lightgreen}0.8679 & \cellcolor{lightgreen}0.8329 & 0.8315 & \cellcolor{lightgreen}0.7952 & \cellcolor{lightgreen}\textbf{0.8056} & \cellcolor{lightgreen}0.8445 & \cellcolor{lightgreen}0.8545 \\
& & Mistral-7b 
& \cellcolor{lightgreen}\textbf{0.7816} & \cellcolor{lightgreen}\textbf{0.7805} & \cellcolor{lightgreen}0.8611 & \cellcolor{lightgreen}0.8753 & \cellcolor{lightgreen}0.8451 & 0.8424 & \cellcolor{lightgreen}\textbf{0.8088} & \cellcolor{lightgreen}0.7957 & \cellcolor{lightgreen}\textbf{0.8513} & \cellcolor{lightgreen}0.8647 \\
\hline
\end{tabular}
\end{table*}

Additionally, we use a supervised learning task -- \textbf{next-token prediction} -- to fine-tune our model. The LLM receives $\mathbf{X}_{\text{LLM}}$ as input and produces two possible types of output: (i) for classification tasks (link prediction and edge classification), the LLM performs next-token prediction, outputting discrete labels such as '1' or '0' for link prediction, or a categorical token like '[label\_class]' for edge classification;
(ii) for generative tasks (e.g., edge text generation), the LLM produces a full textual sequence that describes the edge. Then, the next token prediction loss is defined as:
\begin{equation}
\mathcal{L}_{\text{LLM}} = \text{CrossEntropy}(\mathbf{o}_{p-1}, \mathbf{x}_p)
\end{equation}
where \( \mathbf{o}_{p-1} \) are the logits at position \( p-1 \), and \( \mathbf{x}_p \) is the ground truth next token, representing the edge label token. We jointly optimize both structural and textual objectives:
\begin{equation}
\label{loss}
\mathcal{L} = \lambda_1 \mathcal{L}_{\text{edge}} + \lambda_2 \mathcal{L}_{\text{LLM}}
\end{equation}
\( \lambda_1 \) and \( \lambda_2 \) control the relative importance of structural and textual supervision. This unified architecture enables flexible edge prediction by jointly modeling structural dynamics and semantic context from the evolving graph. For \textit{text generation} task,  where the goal is to generate edge descriptions via next-token prediction, we rely solely on the language modeling loss by setting \( \lambda_1 = 0 \) and \( \lambda_2 = 1 \), effectively optimizing only \( \mathcal{L}_{\text{LLM}} \).
The impact of different $\lambda_1$ and $\lambda_2$ configurations on model performance is systematically evaluated in the ablation study (see Section~\ref{sec:ablation}).


\textbf{Inference time.} For  \textit{link prediction}, a binary decision is made by thresholding the score \( \hat{y} \). For  \textit{edge classification}, the predicted label is obtained via \( \arg\max \hat{y} \) over the labels output logits, and for \textit{text generation}, we apply standard LLM decoding to autoregressively generate the edge text. 


\textbf{Scaling}. The LLM has a fixed context window, which bounds how much raw text or history can be provided per example. In contrast, the temporal GNN can incorporate long-range history over the evolving graph via minibatched temporal neighbor sampling, producing compact node states that summarize structure and temporal context. We inject these projected TGNN vectors alongside a short textual prompt \emph{without increasing the sequence length}, so compute and memory are dominated by the sampled neighborhood and LLM context length—enabling us to scale to \textbf{large-scale graphs}. Even when full histories are too long to fit and must be downsampled to match the LLM context, the TGNN still provides complementary long-range signals that the LLM alone cannot capture. With adapter-based tuning (QLoRA), the backbone remains frozen and the number of trainable parameters is small, making the framework practical and scalable. Computational efficiency results (wall-clock training time on Google dataset) are reported in Appendix~\ref{app:efficiency_google}.



\subsubsection{Negative Sampling for the Link Prediction Task}

Temporal graph datasets typically include only observed interactions—i.e., positive edges $(u, v)$ at time $t_{u,v}$—with no explicit negative examples. For the \textit{temporal link prediction} task, models must learn to distinguish real interactions from plausible but unobserved ones. This necessitates the generation of negative samples.
Following \cite{zhang2024dtgb}, we adopt a temporal negative sampling strategy. For each positive edge $(u, v)$ at time $t_{u,v}$, we sample a negative edge $(u, v')$, where:
$v' \in V_T \setminus \mathcal{N}_u(t_{u,v})$, $\mathcal{N}_u(t_{u,v})$ denotes the set of all nodes with which node $u$ has interacted prior to timestamp $t_{u,v}$. This guarantees that $v'$ is a node that $u$ 
has not interacted with up to time $t_{u,v}$, preserving the temporal validity of the negative sample. 
To ensure temporal consistency, the sampled $v'$ is chosen from the same temporal window as the original interaction. This process results in a hard negative example $(u, v', t_{u,v})$ that is structurally and temporally plausible yet absent from the interaction history.


\begin{table*}[t]
\caption{Edge Classification results across datasets. F1(W) denotes the weighted F1 score, while F1(M) represents the macro-averaged F1 score across categories.}
\label{edge_classification_table}
\centering
\scriptsize
\setlength{\tabcolsep}{0.9pt}
\renewcommand{\arraystretch}{1.4}
\begin{tabular}{|c|c|c|ccc|ccc|ccc|ccc|ccc|}
\hline
\textbf{Method} & \parbox[c]{1cm}{\centering \textbf{Graph}\\\textbf{Models}} & \textbf{LLM}
& \multicolumn{3}{c|}{\textbf{Enron}} 
& \multicolumn{3}{c|}{\textbf{Googlemap CT}} 
& \multicolumn{3}{c|}{\textbf{Stack Elec}} 
& \multicolumn{3}{c|}{\textbf{GDELT}} 
& \multicolumn{3}{c|}{\textbf{Amazon Movies}} \\
\cline{4-18}
 & & & \textbf{F1 (W)} & \textbf{Acc.} & \textbf{F1 (M)} 
 & \textbf{F1 (W)} & \textbf{Acc.} & \textbf{F1 (M)} 
 & \textbf{F1 (W)} & \textbf{Acc.} & \textbf{F1 (M)} 
 & \textbf{F1 (W)} & \textbf{Acc.} & \textbf{F1 (M)} 
 & \textbf{F1 (W)} & \textbf{Acc.} & \textbf{F1 (M)}  \\
\hline

\multirow{3}{*}{\parbox{1.1cm}{\centering \textbf{Graph-based}\\\textbf{methods}}}
& DyGFormer & -
& 0.4180 & 0.4288 & 0.2404 & 0.5816 & 0.6684 & 0.2444 & 0.6615 & 0.7556 & \cellcolor{lightyellow}0.4438 & 0.0684 & 0.1140 & 0.0056 & 0.5654 & 0.6711 & 0.2111 \\
& GraphMixer & -
& 0.4209 & 0.4338 & 0.2324 & 0.5695 & 0.6691 & 0.2202 & 0.6602 & 0.7567 & 0.4453 & 0.0759 & 0.1192 & 0.0067 & 0.5615 & 0.6719 & 0.1827 \\
& TGAT & -
& 0.3397 & 0.3840 & 0.1441 & 0.5600 & 0.6695 & 0.2077 & 0.6545 & 0.7579 & 0.4337 & 0.0765 & 0.1128 & 0.0074 & 0.5669 & 0.6720 & 0.1972 \\
\hline
\hline

\multirow{3}{*}{\parbox{1.1cm}{\centering \textbf{LLM-based}\\\textbf{methods}}}
& - & Llama3-1B 
& 0.5563 & 0.5496 & 0.1364 & 0.5938 & 0.6357 & 0.2648 & 0.6550 & 0.7298 & 0.0171 & 0.0588 & 0.0746 & 0.0068 & 0.5701 & 0.6438 & 0.2127 \\
& - & Llama3-3B 
& 0.5625 & 0.5518 & 0.4134 & 0.5953 & 0.6437 &  \cellcolor{lightyellow} 0.2727 & 0.6523 & 0.7452 & 0.4056 & 0.0655 & 0.0805 & \cellcolor{lightyellow}0.0110 & 0.5776 & 0.6487 & 0.2109 \\
& - & Mistral-7b 
& 0.5637 & 0.5349 & 0.0748 & 0.6167 & 0.6784 & 0.1111 & 0.6506 & 0.6935 & 0.0646 & 0.0625 & 0.1167 & 0.0109 & 0.3769 & 0.3253 & 0.1523 \\
\hline
\hline

\multirow{9}{*}{\parbox{1.1cm}{\centering \textbf{Ours –} \\\textbf{Temporal}\\\textbf{Graph LLM}}}
& DyGFormer & Llama3-1B 
&\cellcolor{lightgreen}\textbf{0.6133} & \cellcolor{lightgreen}\textbf{0.6069} & \cellcolor{lightgreen}\textbf{0.4696} & \cellcolor{lightgreen} 0.6159 & \cellcolor{lightgreen}0.6835 & \cellcolor{lightgreen} 0.2798 & 0.6540 & \cellcolor{lightgreen}0.7578 & 0.4327 & \cellcolor{lightgreen}0.0814 & \cellcolor{lightgreen}0.1215 & \cellcolor{lightgreen}0.0086 & \cellcolor{lightgreen}0.5731 & \cellcolor{lightgreen}0.6787 & 0.2122 \\
& & Llama3-3B 
& \cellcolor{lightgreen}0.5817 &\cellcolor{lightgreen} 0.5823 & \cellcolor{lightgreen}0.4510 &\cellcolor{lightgreen} 0.6047 &\cellcolor{lightgreen} 0.6803 & 0.2664 & 0.6524 & \cellcolor{lightgreen}0.7569 & 0.4317 & \cellcolor{lightgreen}0.0867 &\cellcolor{lightgreen} 0.1272 & 0.0097 & 0.5766 & \cellcolor{lightgreen}0.6790 & \cellcolor{lightgreen}0.2282 \\
& & Mistral-7b 
&\cellcolor{lightgreen} 0.5828 &\cellcolor{lightgreen} 0.5768 & \cellcolor{lightgreen}0.4288 & \cellcolor{lightgreen}\textbf{0.6186} &\cellcolor{lightgreen} 0.6884 & \cellcolor{lightgreen}0.2718 & \cellcolor{lightgreen}0.6699 & \cellcolor{lightgreen}0.7700 & 0.4350 & \cellcolor{lightgreen}0.0907 & \cellcolor{lightgreen}0.1291 & 0.0105 & \cellcolor{lightgreen}0.6048 & \cellcolor{lightgreen}\textbf{0.6838} &\cellcolor{lightgreen} 0.2717 \\

& GraphMixer & Llama3-1B 
& \cellcolor{lightgreen} 0.5937 & \cellcolor{lightgreen} 0.5801 &\cellcolor{lightgreen} 0.4492 & \cellcolor{lightgreen} 0.6140 & \cellcolor{lightgreen}0.6830 & \cellcolor{lightgreen}0.2767 &\cellcolor{lightgreen} 0.6602 & \cellcolor{lightgreen}0.7583 & \cellcolor{lightgreen}0.4458 &\cellcolor{lightgreen} 0.0811 &\cellcolor{lightgreen} 0.1236 & \cellcolor{lightgreen}0.0082 & \cellcolor{lightgreen}0.5756 & \cellcolor{lightgreen}0.6777 &\cellcolor{lightgreen} 0.2158 \\
& & Llama3-3B 
& \cellcolor{lightgreen}0.5812 & \cellcolor{lightgreen}0.5840 &\cellcolor{lightgreen} 0.4298 & \cellcolor{lightgreen} 0.6053 &\cellcolor{lightgreen} 0.6792 & 0.2574 & \cellcolor{lightgreen}0.6687 & \cellcolor{lightgreen}0.7572 & \cellcolor{lightgreen}\textbf{0.4664} & \cellcolor{lightgreen}0.0901 & \cellcolor{lightgreen}0.1305 & 0.0102 & \cellcolor{lightgreen}0.5874 & \cellcolor{lightgreen}0.6775 & \cellcolor{lightgreen}0.2417 \\
& & Mistral-7b 
& \cellcolor{lightgreen}0.5802 & \cellcolor{lightgreen}0.5766 & \cellcolor{lightgreen}0.4245 &\cellcolor{lightgreen}0.6183 & \cellcolor{lightgreen}\textbf{0.6888} &\cellcolor{lightgreen} 0.2724 & \cellcolor{lightgreen}\textbf{0.6766} & \cellcolor{lightgreen}\textbf{0.7704} & \cellcolor{lightgreen}0.4498 & \cellcolor{lightgreen}0.0911 &\cellcolor{lightgreen} 0.1291 & \cellcolor{lightgreen}0.0111 & 0.5398 & 0.6714 & 0.1610 \\

& TGAT & Llama3-1B 
& \cellcolor{lightgreen}0.5938 &\cellcolor{lightgreen} 0.5859 & \cellcolor{lightgreen}0.4422 & \cellcolor{lightgreen}0.6161 &\cellcolor{lightgreen} 0.6832 & \cellcolor{lightgreen} \textbf{0.2806} &\cellcolor{lightgreen} 0.6599 & \cellcolor{lightgreen}0.7581 & \cellcolor{lightgreen}0.4452 &\cellcolor{lightgreen} 0.0833 & \cellcolor{lightgreen}0.1240 & \cellcolor{lightgreen}0.0089 & \cellcolor{lightgreen}0.5948 & \cellcolor{lightgreen}0.6803 & \cellcolor{lightgreen}0.2273 \\
& & Llama3-3B 
& \cellcolor{lightgreen}0.5725 &\cellcolor{lightgreen} 0.5776 &\cellcolor{lightgreen} 0.4592 & \cellcolor{lightgreen}0.6046 & \cellcolor{lightgreen}0.6794 & 0.2555 &\cellcolor{lightgreen} 0.6610 & 0.7561 & 0\cellcolor{lightgreen}.4507 & \cellcolor{lightgreen}0.0890 & \cellcolor{lightgreen}0.1295 & 0.0101 & \cellcolor{lightgreen}0.5863 &\cellcolor{lightgreen} 0.6777 & \cellcolor{lightgreen}0.2392 \\
& & Mistral-7b 
& \cellcolor{lightgreen}0.5806 & \cellcolor{lightgreen}0.5802 & \cellcolor{lightgreen}0.4283 & \cellcolor{lightgreen}0.6183 &\cellcolor{lightgreen} 0.6884 & \cellcolor{lightgreen}0.2741 & \cellcolor{lightgreen}0.6742 & \cellcolor{lightgreen}\textbf{0.7704} &\cellcolor{lightgreen} 0.4442 & \cellcolor{lightgreen}\textbf{0.0931} & \cellcolor{lightgreen}\textbf{0.1320} & \cellcolor{lightgreen}\textbf{0.0113} & \cellcolor{lightgreen}\textbf{0.6118} & \cellcolor{lightgreen}0.6812 & \cellcolor{lightgreen}\textbf{0.2793} \\
\hline
\end{tabular}
\end{table*}

\section{Experimentation}

\subsection{Datasets} \label{datasets}
We used the datasets provided in \cite{zhang2024dtgb}\footnote{https://github.com/zjs123/DTGB.}, where nodes represent entities such as users, products, questions, and edges capture relationships like transactions and reviews. Edge categories represent predefined criteria that align with real-world scenarios, such as product ratings and content topics. Dataset statistics are shown in Table \ref{tab:dataset_stats}. The description of each dataset is detailed in Appendix \ref{appendix_datasets}.
We follow DyGLib \cite{yu2023towards} for dataset handling, training, and evaluation. 
\subsection{Tasks}
\textbf{Future Link Prediction.} This task evaluates whether two nodes will form a connection at timestamp \( T + 1 \), given the entire graph history up to time \( T \) \cite{tgat2019,graphmixer2023,dygformer2023}. In dynamic text-attributed graphs, predictions rely on both the evolving structure and rich textual context from nodes and edges. We use an inductive evaluation setup: the model is asked to predict links that may involve new nodes not seen during training, reflecting real-world scenarios such as forecasting future communications (e.g., email exchanges) involving both known and previously unseen participants. This setup tests the model’s ability to generalize to new entities and evolving interactions.

\textbf{Edge Classification.} Given historical edges up to time \( T \), the goal is to predict the category of a future interaction at \( T + 1 \). The model leverages node/edge textual attributes and temporal structure. Although underexplored in prior work \cite{dygformer2023, huang2023temporal}, this task is key for applications like review rating prediction.

\textbf{Textual Relation Generation.} This task involves generating the natural language text of a future interaction between nodes \( u \) and \( v \), using prior interactions and their textual content. Unlike typical TAG tasks \cite{tags,graphllm,tang2024graphgpt} that predict structure, this task evaluates a model’s ability to jointly model dynamic topology and evolving language via LLM-TGNN fusion.


\subsection{Baselines}
We use the top-performing TGNN baselines from DTGB~\cite{zhang2024dtgb}: TGAT~\cite{tgat2019}, GraphMixer~\cite{graphmixer2023}, and DyGFormer~\cite{dygformer2023}.

We compare TemporalGraphLLM to several state-of-the-art baselines from both the Temporal GNN and LLM domains. This comparison highlights the performance of each model individually and demonstrates how the combination of both within our framework outperforms existing approaches.
For \textbf{Temporal Graph Models}, we experiment with the top-performing methods identified in \cite{zhang2024dtgb} as baselines for link prediction and edge classification: TGAT~\cite{tgat2019}, GraphMixer~\cite{graphmixer2023}, and DyGFormer~\cite{dygformer2023}.




 To ensure equitable comparison, all TGNN baselines use the \emph{same} node text features as our method—fixed Llama-3-1B embeddings of the nodes’ textual descriptions.





\begin{table*}[t]
\caption{Results for edge text generation task. BERT-F1 represents BERTScore F1, SBERT represents Sentence-BERT embedding similarity, and R1 represents ROUGE-1.}
\label{edge_generation_table}
\centering
\scriptsize
\setlength{\tabcolsep}{1pt}
\renewcommand{\arraystretch}{1.4}
\begin{tabular}{|c|c|c|ccc|ccc|ccc|ccc|ccc|}
\hline
\textbf{Method} & \parbox[c]{1cm}{\centering \textbf{Graph}\\\textbf{Model}} & \textbf{LLM}
& \multicolumn{3}{c|}{\textbf{Enron}} 
& \multicolumn{3}{c|}{\textbf{Googlemap CT}} 
& \multicolumn{3}{c|}{\textbf{Stack Elec}} 
& \multicolumn{3}{c|}{\textbf{GDELT}} 
& \multicolumn{3}{c|}{\textbf{Amazon Movies}} \\
\cline{4-18}
 & & & \textbf{BERT-F1} & \textbf{SBERT} & \textbf{R1} 
 & \textbf{BERT-F1} & \textbf{SBERT} & \textbf{R1} 
 & \textbf{BERT-F1} & \textbf{SBERT} & \textbf{R1} 
 & \textbf{BERT-F1} & \textbf{SBERT} & \textbf{R1} 
 & \textbf{BERT-F1} & \textbf{SBERT} & \textbf{R1}  \\
\hline

\multirow{3}{*}{\parbox{1.1cm}{\centering \textbf{LLM-based}\\\textbf{methods}}}
& - & Llama3-1B     & 0.7801 & 0.2192 & 0.1046 & 0.8483 & 0.3123 & 0.1044 & 0.8396 & 0.2199 & 0.1416 & 0.8630& 0.2573 & 0.1604 &0.8569 & 0.3665 & 0.1449 \\
& - & Llama3-3B     & 0.8001 & 0.1783 & 0.075 & 0.8159 & 0.1596 & 0.0213 & \cellcolor{lightyellow}0.8048 & 0.0394 & 0.0182 & 0.8621 & 0.2564 & 0.1619 & 0.8464 & 0.2006 & 0.0459 \\
& - & Mistral-7B    & 0.7789 & 0.2742 & 0.1448 & 0.8446 & 0.3245 & 0.1069 & 0.5275 & 0.1186 & 0.0622 & 0.8645& 0.2596 & 0.1793 & 0.6588 & 0.1873 & 0.0354 \\
\hline
\hline

\multirow{9}{*}{\parbox{1.1cm}{\centering \textbf{Ours –} \\\textbf{Temporal}\\\textbf{Graph LLM}}}
& DyGFormer & Llama3-1B  & \cellcolor{lightgreen}0.8335 & \cellcolor{lightgreen}0.3811 & \cellcolor{lightgreen}0.2318 &  \cellcolor{lightgreen}0.8521 &  \cellcolor{lightgreen}0.3244 &  \cellcolor{lightgreen}0.1087 & 0.8396 & 0.2199 & \cellcolor{lightgreen} \textbf{0.1420} & \cellcolor{lightgreen}0.8639 & \cellcolor{lightgreen}0.2609 & \cellcolor{lightgreen}0.1629 & \cellcolor{lightgreen} \textbf{0.8674} & \cellcolor{lightgreen} \textbf{0.3743} & \cellcolor{lightgreen} \textbf{0.1528} \\
&           & Llama3-3B  & \cellcolor{lightgreen}\textbf{0.8481} & \cellcolor{lightgreen}\textbf{0.4409} &\cellcolor{lightgreen} \textbf{0.2860} & \cellcolor{lightgreen} 0.8259 & 0.1592 & 0.0211 & 0.8020 & \cellcolor{lightgreen} 0.0403 & \cellcolor{lightgreen} 0.0186 & \cellcolor{lightgreen}0.8632 & \cellcolor{lightgreen}0.2630 & \cellcolor{lightgreen}0.1622 & 0.8419 & 0.1613 & 0.0230 \\
&           & Mistral-7B &\cellcolor{lightgreen} 0.7965 & \cellcolor{lightgreen}0.3801 & \cellcolor{lightgreen}0.2205 & 0.8421 & 0.3098 & 0.0967 & \cellcolor{lightgreen}0.7743 & \cellcolor{lightgreen}0.1488 &\cellcolor{lightgreen} 0.0840 & \cellcolor{lightgreen} \textbf{0.8678} & \cellcolor{lightgreen} \textbf{0.2691} & \cellcolor{lightgreen} \textbf{0.1919} &  \cellcolor{lightgreen}0.7393 & \cellcolor{lightgreen}  0.2022 & 0.0325 \\

& GraphMixer & Llama3-1B  & \cellcolor{lightgreen}0.7885 &\cellcolor{lightgreen} 0.3646 & \cellcolor{lightgreen}0.2189 &  \cellcolor{lightgreen}0.8542 &  \cellcolor{lightgreen}0.3459 &  \cellcolor{lightgreen}0.1234 & \cellcolor{lightgreen} \textbf{0.8402} & \cellcolor{lightgreen} \textbf{0.2202} & 0.1412 & 0.8629 & \cellcolor{lightgreen}0.2624 & \cellcolor{lightgreen}0.1645 & 0.8541 & 0.2503 & 0.0735 \\
&            & Llama3-3B  & \cellcolor{lightgreen}0.8022 & \cellcolor{lightgreen}0.1963 & \cellcolor{lightgreen}0.0922 & \cellcolor{lightgreen} 0.8302 & \cellcolor{lightgreen} 0.1909 & \cellcolor{lightgreen} 0.0363 & 0.6508 & \cellcolor{lightgreen} 0.0606 & \cellcolor{lightgreen} 0.0256 & 0.8629 & \cellcolor{lightgreen}0.2600 & \cellcolor{lightgreen}0.1633 & 0.8427 & 0.1670 & 0.0258\\
&            & Mistral-7B & \cellcolor{lightgreen}0.8270 & \cellcolor{lightgreen}0.4145 & \cellcolor{lightgreen}0.2529 & \cellcolor{lightgreen} 0.8484 & \cellcolor{lightgreen}0.3354 & \cellcolor{lightgreen}0.1135 & \cellcolor{lightgreen} 0.6812 & \cellcolor{lightgreen} 0.1390 & \cellcolor{lightgreen} 0.0756 &\cellcolor{lightgreen} 0.8676 &\cellcolor{lightgreen} 0.2680 &\cellcolor{lightgreen} 0.1910 & 0.6523 &\cellcolor{lightgreen} 0.1895 & 0.0337 \\

& TGAT & Llama3-1B        & \cellcolor{lightgreen}0.8056 & \cellcolor{lightgreen}0.2201 & \cellcolor{lightgreen}0.1049 &  \cellcolor{lightgreen}\textbf{0.8561} &  \cellcolor{lightgreen}\textbf{0.3598} &  \cellcolor{lightgreen}\textbf{0.1299} & 0.8391 & 0.2088 & 0.1359 &\cellcolor{lightgreen} 0.8635 & \cellcolor{lightgreen}0.2610 & \cellcolor{lightgreen}0.1646 & \cellcolor{lightgreen}0.8673 & 0.3638 & \cellcolor{lightgreen} 0.1458 \\
&      & Llama3-3B        & \cellcolor{lightgreen}0.8225 &\cellcolor{lightgreen} 0.2912 & \cellcolor{lightgreen}0.1685 & \cellcolor{lightgreen}0.8449 &\cellcolor{lightgreen} 0.2936 & \cellcolor{lightgreen} 0.0891 & 0.6491 & \cellcolor{lightgreen} 0.0489 & \cellcolor{lightgreen} 0.0222 & \cellcolor{lightgreen}0.8626 & \cellcolor{lightgreen}0.2583 & 0.1597 & \cellcolor{lightgreen}0.8488 &\cellcolor{lightgreen} 0.2114 & \cellcolor{lightgreen}0.0502 \\
&      & Mistral-7B       & \cellcolor{lightgreen}0.8320 &\cellcolor{lightgreen} 0.3942 & \cellcolor{lightgreen}0.2479 & 0.8411 & 0.3071 & 0.0950 & \cellcolor{lightgreen}0.7024 &\cellcolor{lightgreen} 0.1619 & \cellcolor{lightgreen}0.0914 & \cellcolor{lightgreen}0.8648 &\cellcolor{lightgreen} 0.2623 & \cellcolor{lightgreen}0.1867 &\cellcolor{lightgreen} 0.7344 & 0.1873 & \cellcolor{lightgreen}0.0365 \\
\hline
\end{tabular}
\end{table*}

Our framework is \textit{model-agnostic}: it can be paired with any LLM, either fine-tuned as a text-only baseline or integrated with our TGNN injection interface.

\textbf{Large Language Model baselines.}
We fine-tune instruction-tuned LLMs using temporal prompts constructed from historical, time-stamped interactions across all three tasks. For edge classification, the model predicts the next token corresponding to the edge category. For link prediction, the model outputs a binary decision (0/1) as the next token. For textual relation generation, the model generates the full interaction text. Full fine-tuning details are provided in Appendix~\ref{setup}. We evaluate the following open-weight models:
\textbf{Mistral (7B)} \cite{mistral2023mistral7b} is a decoder-only transformer optimized for efficient inference with grouped-query and sliding-window attention. We use Mistral-7B-Instruct-v0.3.

\textbf{Llama-3} \cite{dubey2024Llama} is a family of instruction-tuned language models trained via supervised fine-tuning and RLHF. We use two variants: Llama-3.2-1B-Instruct and Llama-3.2-3B-Instruct.

We focus on open-weight LLMs to ensure controlled fine-tuning and reproducible evaluation; API-only models are excluded since they cannot be matched under the same training protocol and would only be comparable in zero-shot mode, which often underperforms in our setting.

TemporalGraphLLM was trained with combinations of the models specified in the Temporal Graph Models and Large Language Models sections.

\subsection{Metrics}
Evaluating DTAG models requires metrics for structural prediction \emph{and} semantic generation.
For link prediction we report \textbf{Accuracy} and \textbf{F1} as primary metrics, following prior LLM baselines that emit \emph{discrete} labels rather than calibrated probabilities.
To ensure comparability, we select a single decision threshold on the validation set (accuracy optimal) and apply it unchanged to the test set.
For edge classification (multi-class), we report Accuracy and F1 (Macro/Weighted).
For edge text generation, we employ three complementary metrics: (1) ROUGE-1\cite{lin-2004-rouge}: Measures unigram overlap between generated and reference texts; (2) BERTScore F1\cite{zhang2019bertscore}: Uses BERT's contextual embeddings to assess token-level semantic similarity; (3) SBERT Similarity \cite{reimers2019sentence}: Computes cosine similarity between Sentence-BERT embeddings to capture sentence-level semantic alignment.

\subsection{Results}
\label{sec:results}
Tables~\ref{link_prediction_table}--\ref{edge_generation_table} report results for link prediction, edge classification, and edge text generation (setup in Appendix~\ref{setup}). Across all datasets, TGNN backbones (TGAT, GraphMixer, DyGFormer), and LLM sizes (1B/3B/7B), \textbf{TemporalGraphLLM consistently outperforms both TGNN-only and LLM-only baselines} on link prediction and edge classification, demonstrating that injecting temporally-aware TGNN states into the LLM provides complementary structural--temporal signals beyond text prompting alone. 
For edge text generation, improvements are \emph{smaller but consistent}, since generation is dominated by language modeling and temporal prompts already provide strong contextual grounding for LLM-only baselines (see Appendix~\ref{edge_generation_examples}). 
Nevertheless, TGNN representations add long-range temporal structure that cannot be fully captured within the LLM context window. 

Best scores per dataset-task are shown in \textbf{bold};  \colorbox{lightgreen}{green} cells indicate configurations where fusion improves over both corresponding TGNN-only and LLM-only variants, while rare dataset-specific cases where a baseline matches/exceeds all related fusion variants are marked in light \colorbox{lightyellow}{yellow}. 

Statistical testing confirms robustness: Kruskal--Wallis shows significant differences among methods (p$<0.01$) for link prediction and edge classification, and paired Wilcoxon post-hoc tests show the best TemporalGraphLLM variant significantly outperforms all baselines (p$<0.05$) on both tasks. These findings establish TemporalGraphLLM as a state-of-the-art method for learning in dynamic, text-attributed graphs.

\subsection{Analysis and Ablation Study}
\label{sec:ablation}
We ablate key design choices on Enron and Googlemap CT using the Llama3-1B backbone, and report edge classification (F1-Weighted), as it best reflects the TGNN--LLM fusion. We include TGAT, GraphMixer, and DyGFormer to verify that trends hold across TGNN architectures.

\textbf{Number of temporal neighbors ($k$).}
We vary the number of sampled temporal neighbors injected into the prompt, $k \in \{0,3,5,10,15\}$.
As shown in Figure~\ref{fig:k_ablation} and Table~\ref{tab:ablation_k} in Appendix~\ref{ablation_appendix}, performance improves up to $k=10$ and then saturates; using no neighbors ($k=0$) causes a large drop, confirming that temporal neighborhood context is critical. We therefore use $k=10$ by default.

\textbf{Loss weighting ($\lambda_1,\lambda_2$).}
We evaluate different balances between structural supervision ($L_{\text{edge}}$) and next-token supervision ($L_{\text{LLM}}$) in Eq.~\ref{loss}.
Specifically, we test $(\lambda_1,\lambda_2)\in\{(1,1), (1,0), (0,1), (1,2), (2,1)\}$ to evaluate how varying the relative contribution of structural versus textual learning affects model performance.
As shown in Figure~\ref{fig:lambda_ablation} and Table~\ref{tab:ablation_lambda} in Appendix~\ref{ablation_appendix}, performance is stable across a wide range of settings, with a slight advantage for emphasizing structural supervision ($(\lambda_1,\lambda_2)=(2,1)$). Purely textual $(0,1)$ and structure-only $(1,0)$ variants consistently underperform, confirming the benefit of fusing both objectives.
Full detailed analysis is provided in the Appendix.

\section{Conclusions and Future Work}
We presented \textbf{TemporalGraphLLM}, a framework that integrates TGNNs with LLMs for dynamic, text-attributed graph tasks. By injecting temporally-aware node embeddings into the LLM token stream, our approach fuses structural and textual information, consistently outperforming GNN-only and LLM-only baselines across diverse graph domains.
Future directions include extending TemporalGraphLLM to jointly predict edge descriptions and labels, enabling the generation of reasoning sequences that capture temporal and relational context prior to classification, thereby enhancing both interpretability and accuracy. Furthermore, incorporating reinforcement learning could optimize temporal path selection by rewarding reasoning trajectories and generated edge descriptions that lead to correct predictions, promoting more robust and context-aware graph reasoning.


\bibliography{references}
\bibliographystyle{icml2026}

\newpage
\appendix
\onecolumn
\section{Prompt Templates} \label{prompts}
\begin{figure}
    \centering
    \includegraphics[width=0.4\textwidth]{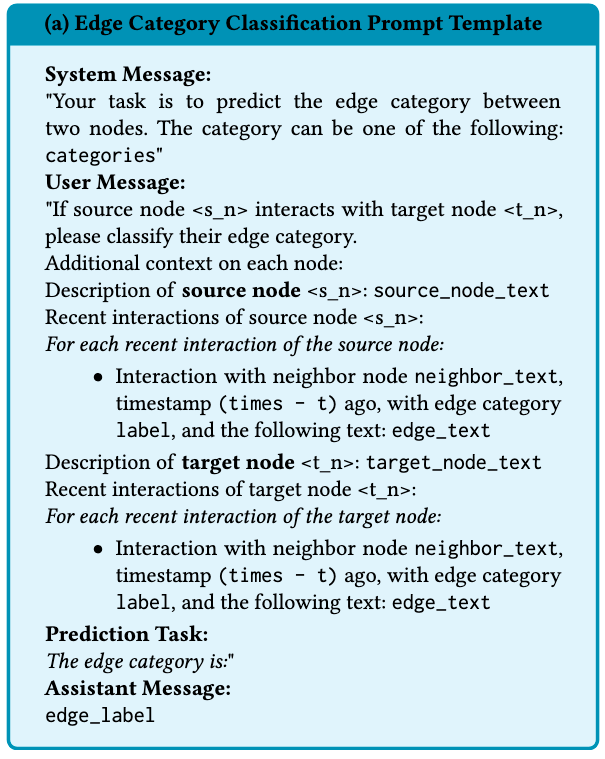}
    \caption{Edge Category Classification Prompt Template}
    \label{fig:prompt_a}
\end{figure}

\begin{figure}
    \centering
    \includegraphics[width=0.4\textwidth]{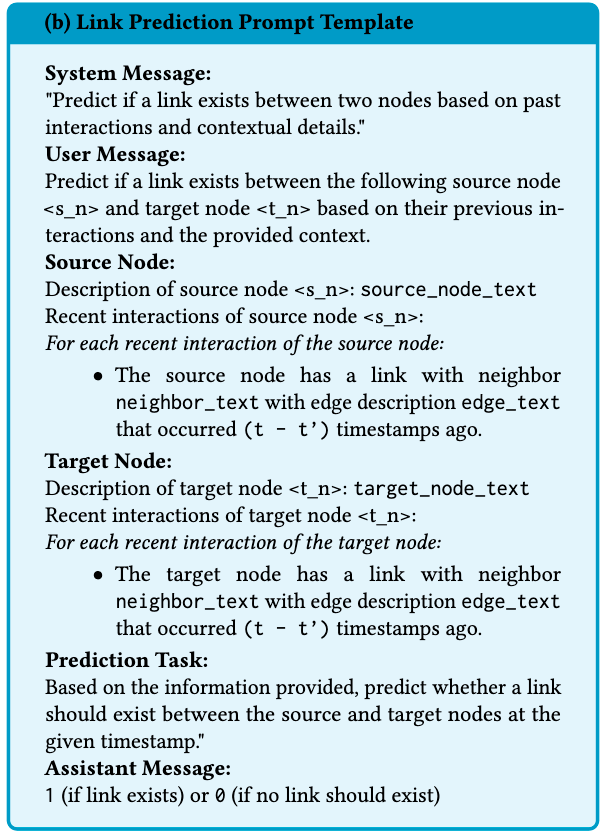}
    \caption{Link Prediction Prompt Template}
    \label{fig:prompt_b}
\end{figure}

\begin{figure}
    \centering
    \includegraphics[width=0.4\textwidth]{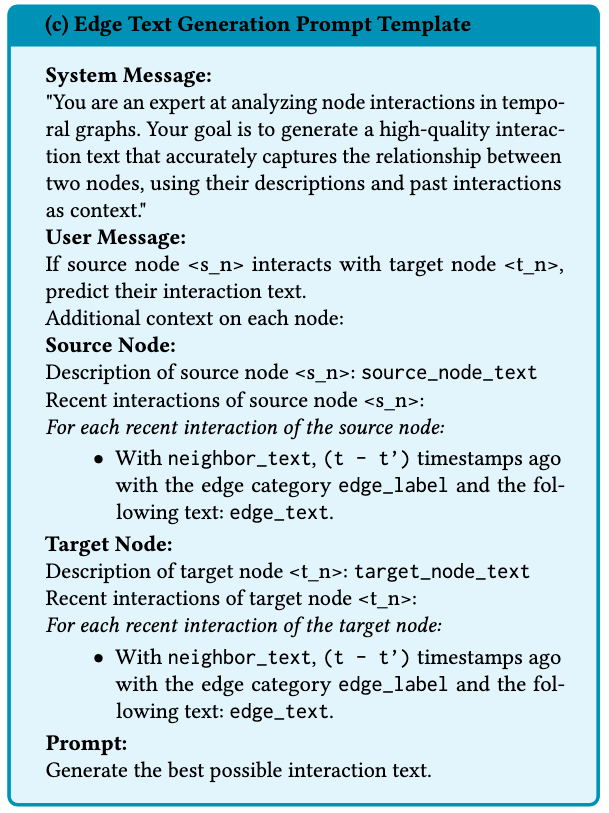}
    \caption{Edge Text Generation Prompt Template}
    \label{fig:prompt_c}
\end{figure}

We employ three prompt templates designed for distinct downstream tasks (Figures~\ref{fig:prompt_a},~\ref{fig:prompt_b},~\ref{fig:prompt_c}): (a) \textbf{Edge Category Classification}, (b) \textbf{Link Prediction}, and (c) \textbf{Edge Text Generation}. The templates incorporate node descriptions, recent temporal interactions, and task-specific instructions, ensuring consistency across classification, prediction, and generative objectives.

\section{Datasets Details} \label{appendix_datasets}

\textbf{Enron}. This dataset consists of email communications within the ENRON corporation (1999-2002). Nodes represent employees, and edges are emails between them. Text attributes of nodes are derived from employees' departments and positions, while edges have raw email text. The dataset has 10 edge categories based on email content, with edges ordered by sending timestamps.

\textbf{GDELT4}. Derived from the Global Database of Events, Language, and Tone project, this dataset tracks global political events. Nodes are political entities (e.g., countries, leaders), and edges represent relationships (e.g., "President of"). Edge text attributes describe the relationships, and edges are ordered by event timestamps.


\textbf{Stack elec}. This dataset is based on Stack Exchange data related to electronics, with nodes representing users and questions. Edges are the answers and comments from users to questions. Text attributes include self-introductions for users and question titles/bodies, while edges have raw answer/comment text and are categorized by usefulness, ordered by answering timestamps.


\textbf{Googlemap CT}. Extracted from the Google Local Data project, this dataset contains reviews of businesses in Connecticut. Nodes represent users and businesses, with text attributes for business information. Edges are user reviews with ratings (1 to 5), and edges are ordered by review timestamps.

\textbf{Amazon movies}. Derived from Amazon Review Data, this dataset includes product reviews for movies and TV. Nodes represent products, and edges are reviews from users, with ratings (1 to 5). Edges are ordered by review timestamps.

\section{Experimentation Setup} \label{setup}
We split each dataset chronologically into train/validation/test sets with a ratio of 7:1.5:1.5. The experiments are conducted on an NVIDIA A40 GPU with 48GB of memory. 
For both TemporalGraphLLM and LLM-based methods, we designed a temporal-oriented prompt (see Appendix~\ref{prompts}). We dynamically inject the relevant historical interaction information into the prompt, using temporal neighbor sampling with $k{=}10$ neighbors, as motivated by the ablation study (Section~\ref{sec:ablation}). 
We used an LLM context length of 2,048 tokens. The same train, validation, and test datasets were used across all models for consistency (including Temporal Graph Models).

For the fine-tuning process, the base model is loaded with LoRA adapters \cite{hu2021lora}. Specifically, the rank of LoRA (\(r\)) is set to 16, and the LoRA scaling factor (\(\alpha\)) is also set to 16. The LoRA dropout rate is set to 0.05, and no bias is used during the fine-tuning.
We also utilize the BitsAndBytes library for quantization (QLoRA \cite{dettmers2023qlora}. The model is loaded in 4-bit precision with double quantization enabled. 
We use a batch size of 4 and perform gradient accumulation over 16 steps, and a warm-up step count of 100. We train for a single epoch due to the large scale of the dataset, which allows the model to see a substantial number of diverse samples even within one pass. Empirically, we observe that convergence often occurs well before completing a full epoch, as validated by early stopping on a held-out validation set. The learning rate is set to \(5 \times 10^{-5}\), and weight decay is set to 0.01. Mixed-precision training was enabled, and the AdamW optimizer with a cosine learning rate scheduler was used.

All TGNN backbones are trained with $L{=}2$ layers, node embedding output size $d_g{=}768$, a time–encoding vector of size $100$, and dropout $p_{\text{drop}}{=}0.1$. We set number of neighbors for TGNN as $k=20$.
For link prediction and edge classification we set  \( \lambda_1 = 1 \) and \( \lambda_2 = 1 \) and for edge text generation  \( \lambda_1 = 0 \) and \( \lambda_2 = 1 \).
We release code and exact configuration files (QLoRA, TGNN, projector, sampling, and data splits), together with scripts to reproduce all results.

\section{Computational Efficiency: Wall-Clock Training Time (Google)}
\label{app:efficiency_google}

We report in Table ~\ref{tab:runtime_google} end-to-end wall-clock training time on the Google dataset to assess computational efficiency
in temporal learning with substantial textual data. We measure total training time from the first forward pass
to the final checkpoint. The \textsc{LLM-only}
baseline and \textsc{TGNN$\times$LLM} use identical training settings (batch size, sequence length, and optimization hyperparameters); differences therefore reflect the effect of TGNN integration and convergence behavior.

\begin{table}[t]
\centering
\small
\begin{tabular}{lccc}
\toprule
\textbf{LLM backbone} & \textbf{LLM-only} & \textbf{TGNN$\times$LLM} & \textbf{Rel. change} \\
\midrule
Llama-1B   & 8h  & 9h  & +12\% \\
Llama-3B   & 19h & 13h & -32\% \\
Mistral-7B & 28h & 21h & -25\% \\
\bottomrule
\end{tabular}
\caption{End-to-end wall-clock training time on the Google dataset. TGNN integration adds minor overhead for smaller
LLMs but reduces total training time for larger backbones, consistent with faster convergence and more stable optimization.}
\label{tab:runtime_google}
\end{table}

\section{Qualitative Examples of Generated Edge Descriptions}\label{edge_generation_examples}

To illustrate the capabilities of \textbf{TemporalGraphLLM} in generating contextualized edge-level text, we present representative examples of generated edge descriptions across different datasets. These examples reflect the model’s ability to capture semantic, temporal, and relational nuances between interacting nodes. Note that the names mentioned in the generated texts correspond to node identifiers from the original datasets and are used as part of the modeling process.

\begin{itemize}
\item \textbf{Google Maps (Review Text)}
\begin{quote}
\textit{"This place is amazing! The food is delicious and the staff is friendly and attentive. The portions are huge and you can't finish them. The wait time is always minimal and they are always accommodating. I highly recommend this place."}
\end{quote}
This edge text represents a user’s review of a location, capturing sentiment, experience, and service quality.
\item \textbf{Enron (Corporate Communication)}  
\begin{quote}
\textit{"The following is a summary of the information that was discussed during the meeting of the East Power Group. Please feel free to forward any questions to me. gngr 713-853-7751."}  
\end{quote}
\begin{quote}

\textit{"Please see the attached memo re: new Global Counterparty website address. If you have any questions, please call me. Thank you, Simone La."}
\end{quote}

These examples reflect the formal tone and content of internal business communications in the Enron dataset.

\item \textbf{StackOverflow (Technical Discussion)}  
\begin{quote}
\textit{"@MarkRansom Yeah, but I think the 68R is a better choice than a 1k resistor because it's a bit more stable at the same current."}
\end{quote}
The model generates technical and context-aware discussion aligned with typical StackExchange interactions.

\item \textbf{GDELT (Event Classification)}  
\begin{quote}
\textit{"make an appeal or request"}, 
\textit{"make a visit"},  
\textit{"make statement"},
\end{quote}
These outputs reflect concise event descriptions generated for geopolitical interactions, matching predefined edge categories.

\item \textbf{Amazon Movies (User Review)}  
\begin{quote}
\textit{"I'm a big fan of this movie, and I'm glad I could purchase it on Bluray."}
\end{quote}
This description captures typical sentiment and purchasing intent found in product reviews.
\end{itemize}

These qualitative results show how \textbf{TemporalGraphLLM} adapts its generation to different domains, producing relevant, coherent, and context-sensitive edge descriptions.

\section{Ablation Study}\label{ablation_appendix}

\paragraph{\textbf{(1) Temporal vs. Random Neighbor Sampling.}}
To assess the impact of recency-aware sampling, we compare our temporal probability distribution (Eq.~\ref{eq:temporal_sampling}) with uniform random neighbor sampling.
Table~\ref{tab:ablation_sampling} (Enron Dataset) reports results for $k{=}5$ sampled neighbors and shows that temporal sampling consistently outperforms random sampling across all TGNN backbones.
For example, DyGFormer+Llama3-1B improves from \textbf{0.5494} to \textbf{0.5660} in weighted F1 and from \textbf{0.2391} to \textbf{0.4239} in macro F1, confirming that emphasizing recent interactions enhances temporal reasoning.

\begin{figure}[t]
    \centering
    \includegraphics[width=0.6\linewidth]{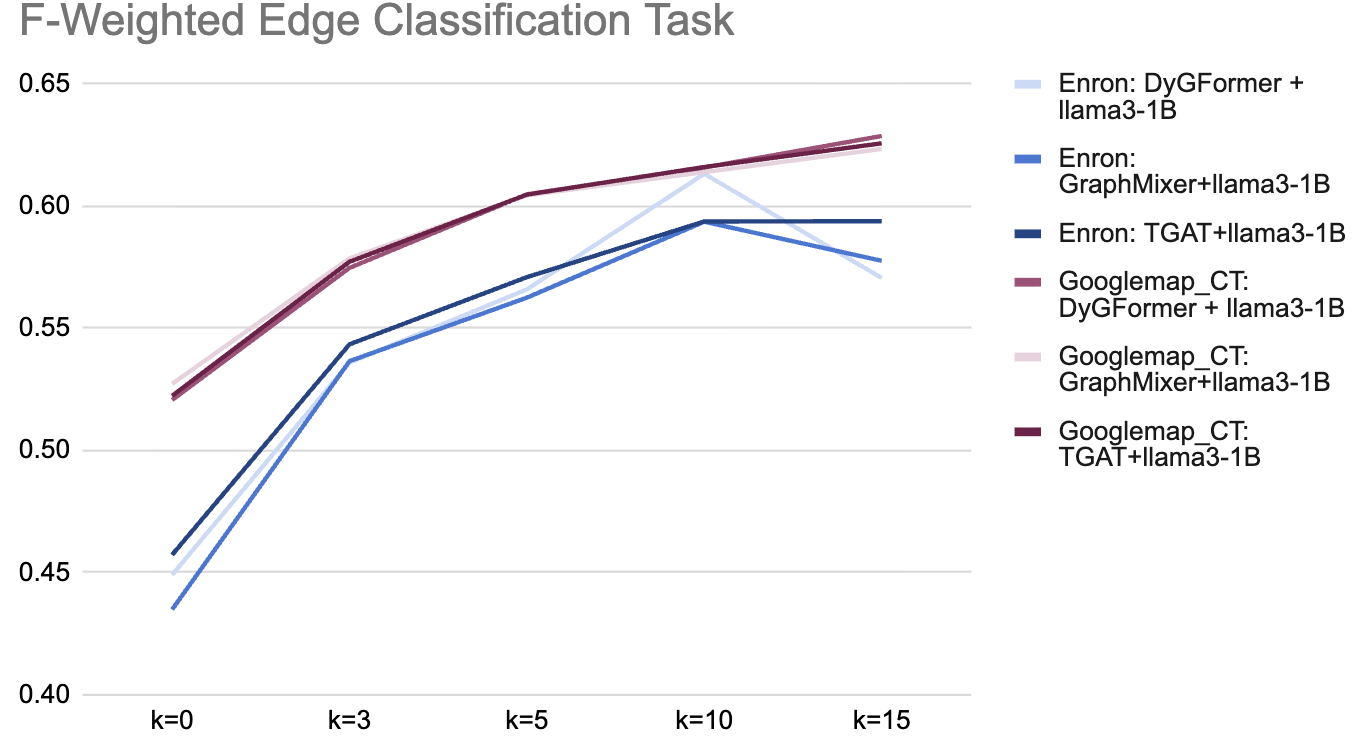}
    \caption{Ablation on the number of sampled temporal neighbors ($k$) for the edge classification task (F1-Weighted metric).
    Performance improves up to $k{=}10$ and plateaus thereafter across both Enron and Googlemap CT datasets.}
    \label{fig:k_ablation}
\end{figure}

\paragraph{\textbf{(2) Effect of the Number of Temporal Neighbors ($k$).}}
Figure ~\ref{fig:k_ablation} and Table~\ref{tab:ablation_k} present the full quantitative results for varying the number of sampled temporal neighbors ($k \in \{0,3,5,10,15\}$)
across all TGNN backbones.
We can see the performance consistently improves up to $k{=}10$ and then stabilizes, indicating that a moderate neighborhood context is sufficient.

\paragraph{\textbf{(3) Effect of Loss Weighting ($\lambda_1,\lambda_2$).}}
We study how different balances between the structural supervision $\mathcal{L}_{\text{edge}}$ and textual supervision $\mathcal{L}_{\text{LLM}}$ affect model performance.
Each configuration uses Llama3-1B with $k{=}10$ sampled temporal neighbors.
The tested setups are $(\lambda_1,\lambda_2) \in \{(1,1),(0,1),(1,0),(1,2),(2,1)\}$.

Even when $\lambda_1{=}0,\lambda_2{=}1$ (text-only supervision), the TGNN still contributes through its injected node embeddings in the LLM’s token space, conditioning the next-token generation on graph context. 
Conversely, when $\lambda_1{=}1,\lambda_2{=}0$ (structure-only supervision), the edge-classification head operates over fused representations that combine TGNN and LLM embeddings, ensuring that textual context remains embedded in the structural signal.
Thus, every configuration reflects a different balance between structural and textual supervision rather than isolating either component.

\begin{figure}[t]
\includegraphics[width=0.6\textwidth]{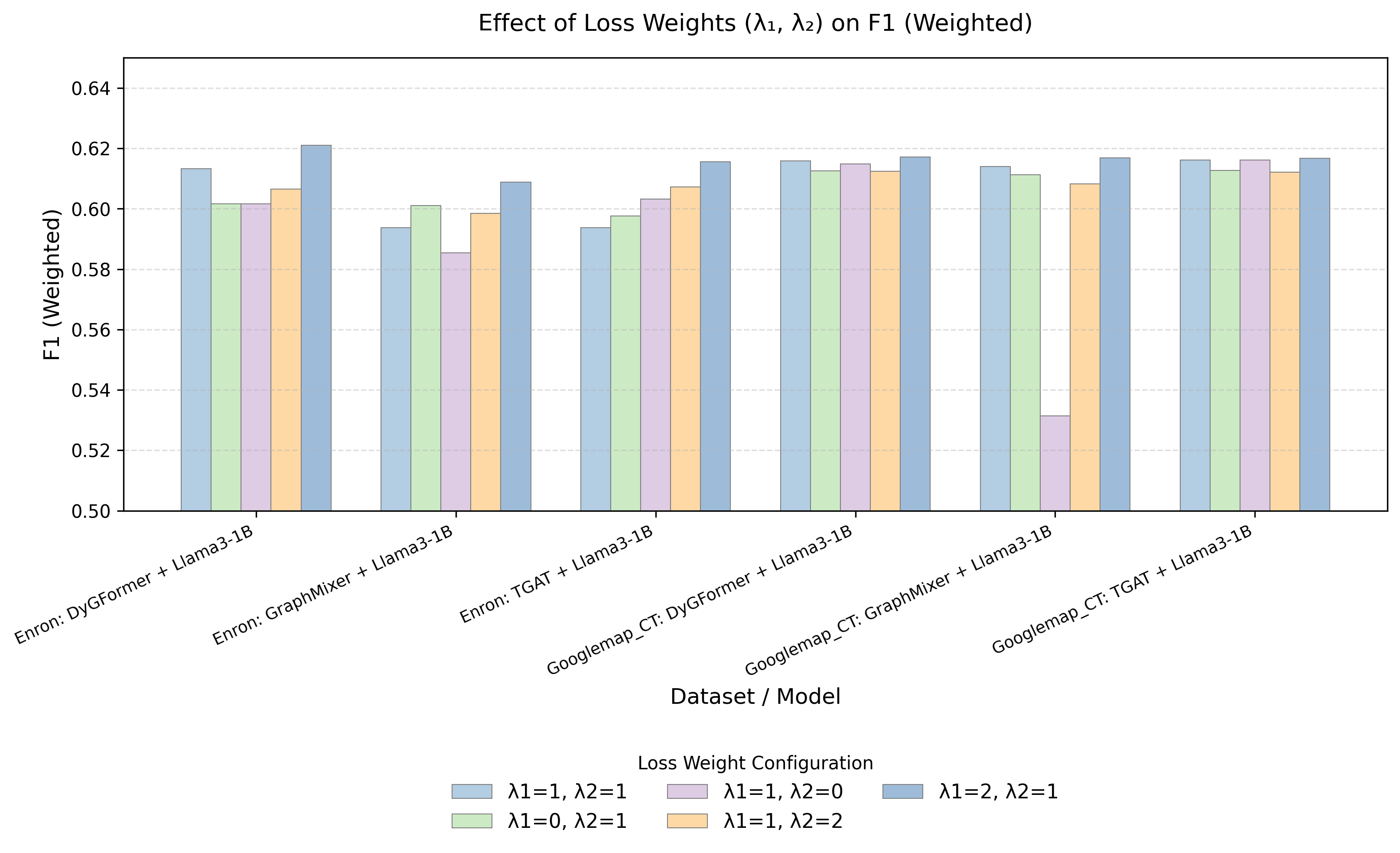}
\centering
\caption{Effect of loss weighting $(\lambda_1,\lambda_2)$ on F1-Weighted for edge classification. 
Each color represents a distinct configuration.}
\label{fig:lambda_ablation}
\end{figure}

Figure ~\ref{fig:lambda_ablation} and Table~\ref{tab:ablation_lambda} report results on both Enron and Googlemap CT datasets across all TGNN backbones.
The patterns are broadly similar between Enron and Googlemap CT:
equal weighting $(1,1)$ provides a strong baseline, and increasing the structural term to $(2,1)$ produces the highest F1(W) scores—e.g., \textit{DyGFormer+Llama3-1B} reaches 0.6211 (Enron) and 0.6172 (Googlemap CT).
This shows that structural supervision remains the dominant driver of accuracy, while textual supervision complements it by stabilizing temporal reasoning.
Purely textual $(0,1)$ and structure-only $(1,0)$ variants consistently underperform, confirming the benefit of fusing both objectives.

\begin{table*}[t]
\centering
\caption{Ablation (1): Temporal vs. Random Neighbor Sampling ($k{=}5$) on Enron with Llama3-1B.
Temporal sampling consistently improves performance across all TGNN backbones.}
\label{tab:ablation_sampling}
\small
\renewcommand{\arraystretch}{1.25}
\setlength{\tabcolsep}{5pt}
\begin{tabular}{l|ccc|ccc}
\hline
\multirow{2}{*}{\textbf{Model}} & \multicolumn{3}{c|}{\textbf{Temporal Neighbor Sampling}} & \multicolumn{3}{c}{\textbf{Random Neighbor Sampling}} \\
 & F1 (Weighted) & Acc & F1 (Macro) & F1 (Weighted) & Acc & F1 (Macro) \\
\hline
DyGFormer + Llama3-1B & \textbf{0.5660} & \textbf{0.5587} & \textbf{0.4239} & 0.5494 & 0.5497 & 0.2391 \\
GraphMixer + Llama3-1B & \textbf{0.5626} & \textbf{0.5574} & \textbf{0.4102} & 0.5615 & 0.5467 & 0.1695 \\
TGAT + Llama3-1B & \textbf{0.5710} & \textbf{0.5653} & \textbf{0.4239} & 0.5623 & 0.5502 & 0.2528 \\
\hline
\end{tabular}
\end{table*}

\begin{table*}[!htbp]
\centering
\caption{Ablation (2): Effect of the Number of Sampled Neighbors ($k$) on Enron and Googlemap CT with Llama3-1B.}
\label{tab:ablation_k}
\scriptsize
\renewcommand{\arraystretch}{1.25}
\setlength{\tabcolsep}{4pt}
\begin{tabular}{l|ccc|ccc|ccc|ccc|ccc}
\hline
\multirow{2}{*}{\textbf{Dataset / Model}} & \multicolumn{3}{c|}{$k{=}0$} & \multicolumn{3}{c|}{$k{=}3$} & \multicolumn{3}{c|}{$k{=}5$} & \multicolumn{3}{c|}{$k{=}10$} & \multicolumn{3}{c}{$k{=}15$} \\
 & F1(W) & Acc & F1(M) & F1(W) & Acc & F1(M) & F1(W) & Acc & F1(M) & F1(W) & Acc & F1(M) & F1(W) & Acc & F1(M) \\
\hline
\multicolumn{16}{c}{\textbf{Enron}} \\
\hline
DyGFormer + Llama3-1B & 0.4491 & 0.4384 & 0.2619 & 0.5360 & 0.5259 & 0.3847 & 0.5660 & 0.5587 & 0.4239 & 0.6133 & 0.6069 & 0.4696 & 0.5707 & 0.5731 & 0.4144 \\
GraphMixer + Llama3-1B & 0.4349 & 0.4246 & 0.2824 & 0.5366 & 0.5308 & 0.3886 & 0.5626 & 0.5574 & 0.4102 & 0.5937 & 0.5801 & 0.4492 & 0.5777 & 0.5729 & 0.4189 \\
TGAT + Llama3-1B & 0.4573 & 0.4499 & 0.2929 & 0.5435 & 0.5364 & 0.3792 & 0.5710 & 0.5653 & 0.4239 & 0.5938 & 0.5859 & 0.4422 & 0.5939 & 0.5860 & 0.4879 \\
\hline
\multicolumn{16}{c}{\textbf{Googlemap CT}} \\
\hline
DyGFormer + Llama3-1B & 0.5207 & 0.6561 & 0.1602 & 0.5748 & 0.6640 & 0.2316 & 0.6049 & 0.6731 & 0.2734 & 0.6159 & 0.6835 & 0.2798 & 0.6288 & 0.7023 & 0.2632 \\
GraphMixer + Llama3-1B & 0.5273 & 0.6566 & 0.1692 & 0.5787 & 0.6658 & 0.2375 & 0.6045 & 0.6714 & 0.2733 & 0.6140 & 0.6830 & 0.2767 & 0.6235 & 0.6996 & 0.2543 \\
TGAT + Llama3-1B & 0.5224 & 0.6570 & 0.1615 & 0.5773 & 0.6656 & 0.2350 & 0.6048 & 0.6706 & 0.2729 & 0.6161 & 0.6832 & 0.2806 & 0.6258 & 0.7002 & 0.2587 \\
\hline
\end{tabular}
\end{table*}

\begin{table*}[ht!]
\centering
\caption{Ablation (3): Effect of Loss Weight Configuration on Enron and Googlemap CT with Llama3-1B. Best F1(W) per dataset in bold.}
\label{tab:ablation_lambda}
\scriptsize
\renewcommand{\arraystretch}{1.25}
\setlength{\tabcolsep}{4pt}
\begin{tabular}{l|ccc|ccc|ccc|ccc|ccc}
\hline
\multirow{2}{*}{\textbf{Dataset / Model}} & \multicolumn{3}{c|}{$\lambda_1{=}1,\lambda_2{=}1$} & \multicolumn{3}{c|}{$\lambda_1{=}0,\lambda_2{=}1$} & \multicolumn{3}{c|}{$\lambda_1{=}1,\lambda_2{=}0$} & \multicolumn{3}{c|}{$\lambda_1{=}1,\lambda_2{=}2$} & \multicolumn{3}{c}{$\lambda_1{=}2,\lambda_2{=}1$} \\
 & F1(W) & Acc & F1(M) & F1(W) & Acc & F1(M) & F1(W) & Acc & F1(M) & F1(W) & Acc & F1(M) & F1(W) & Acc & F1(M) \\
\hline
\multicolumn{16}{c}{\textbf{Enron}} \\
\hline
DyGFormer + Llama3-1B & 0.6133 & 0.6069 & 0.4696 & 0.6017 & 0.5901 & 0.4461 & 0.6017 & 0.5880 & 0.4583 & 0.6065 & 0.5943 & 0.4728 & \textbf{0.6211} & \textbf{0.6098} & \textbf{0.4799} \\
GraphMixer + Llama3-1B & 0.5937 & 0.5801 & 0.4492 & 0.6011 & 0.5865 & 0.4132 & 0.5854 & 0.5741 & 0.4331 & 0.5985 & 0.5907 & 0.4430 & \textbf{0.6088} & \textbf{0.5946} & \textbf{0.4510} \\
TGAT + Llama3-1B & 0.5938 & 0.5859 & 0.4422 & 0.5976 & 0.5883 & 0.4475 & 0.6032 & 0.5912 & 0.4365 & 0.6072 & 0.5927 & 0.4705 & \textbf{0.6156} & \textbf{0.5988} & \textbf{0.4904} \\
\hline
\multicolumn{16}{c}{\textbf{Googlemap CT}} \\
\hline
DyGFormer + Llama3-1B & 0.6159 & 0.6835 & 0.2798 & 0.6125 & 0.6843 & 0.2833 & 0.6148 & 0.6843 & 0.2777 & 0.6124 & 0.6828 & 0.2726 & \textbf{0.6172} & \textbf{0.6849} & \textbf{0.2802} \\
GraphMixer + Llama3-1B & 0.6140 & 0.6830 & 0.2767 & 0.6113 & 0.6838 & 0.2801 & 0.5314 & 0.6632 & 0.1619 & 0.6082 & 0.6802 & 0.2669 & \textbf{0.6169} & \textbf{0.6835} & \textbf{0.2831} \\
TGAT + Llama3-1B & 0.6161 & 0.6832 & 0.2806 & 0.6127 & 0.6844 & 0.2823 & 0.6162 & 0.6836 & 0.2814 & 0.6121 & 0.6822 & 0.2726 & \textbf{0.6167} & \textbf{0.6833} & \textbf{0.2808} \\
\hline
\end{tabular}
\end{table*}
\end{document}